\documentclass{article}
\usepackage{ijcai25}

\usepackage{times}
\usepackage{soul}
\usepackage{url}
\usepackage[hidelinks]{hyperref}
\usepackage[utf8]{inputenc}
\usepackage[small]{caption}
\usepackage{graphicx}
\usepackage{amsmath}
\usepackage{amssymb}
\usepackage{amsthm}
\DeclareMathOperator*{\argmax}{\arg\!\max}

\usepackage{booktabs}
\usepackage{algorithm}
\usepackage{algorithmic}
\usepackage[switch]{lineno}

\newcommand{\citet}[1]{\citeauthor{#1}~[\citeyear{#1}]}

\title{Probing the Embedding Space of Transformers via Minimal Token Perturbations}

\author{
Eddie Conti$^{1,2}$
\and
Alejandro Astruc$^{1,2}$\and
Alvaro Parafita$^{1}$\And
Axel Brando $^1$\\
\affiliations
$^1$ TAIES Group, HPES Lab, Barcelona Supercomputing Center (BSC), Barcelona \\%Department of Computer Sciences, Barcelona Supercomputing Center, Barcelona \\
$^2$Equal Contribution\\
\emails
\{econti, aastruc, aparafita, axel.brando\}@bsc.es,
}

\begin{document}

\maketitle

\begin{abstract}
Understanding how information propagates through Transformer models is a key challenge for interpretability. 
In this work, we study the effects of minimal token perturbations on the embedding space.
In our experiments, we analyze the frequency of which tokens yield to minimal shifts, highlighting that rare tokens usually lead to larger shifts.
Moreover, we study how perturbations propagate across layers, demonstrating that input information is increasingly intermixed in deeper layers.
Our findings validate the common assumption that the first layers of a model can be used as proxies for model explanations. 
Overall, this work introduces the combination of token perturbations and shifts on the embedding space as a powerful tool for model interpretability.

\end{abstract}

\section{Introduction and Related Work}
The widespread use of Transformers \cite{original_paper} in areas like computer vision and text summarization \cite{vision_application,text_application} has led to growing research interest in uncovering the inner workings of this architecture, which remains only partially understood.
According to \citet{survey_expl_transf}, most interpretability techniques for Transformers can be grouped into four categories: (1) \textbf{Activation methods}, which trace the contribution of each neuron to the final output; (2) \textbf{Attention methods}, which use attention weights—especially in early layers—as proxies to estimate the contribution of each token and the relationships among input tokens; (3) \textbf{Gradient methods}, which rely on gradients of the loss function with respect to different components of the network; and (4) \textbf{Perturbation methods}, which assess input relevance by masking parts of the input and measuring the impact on the output.

Among these, attention weights have often been a go-to choice for explanations. However, their validity as explanatory tools remains in question, with contrasting views in the literature \cite{attn_not,attn_not_not}.
The field of explainability is rapidly evolving with new paradigms emerging. \citet{Intepretability_paradigm} discuss the present and emerging paradigms in terms of faithfulness---i.e., the degree to which an explanation resembles the underlying reasoning process of the model---a key aspect in explainability \cite{faithfulness}. In particular, \citet{Intepretability_paradigm} bring attention to the problem of generating out-of-distribution (OOD) instances during the generation of perturbations, hence compromising the validity of explanations that rely on them. In this work, we avoid this problem by performing perturbations aimed at preserving the semantics of the input.

Within the topic of attention methods, identifiability is crucial for interpreting the role of attention in the model: ``attention weights of an attention head for a given input are identifiable if they can be uniquely determined from the head’s output '' \cite{tok_identifiability}. When using attention for model explanations it is implicitly assumed that, at each layer, every position can be associated to the original input token. This assumption is empirically studied by \citet{tok_identifiability} as \textbf{token identifiability} (i.e., the capacity of recovering the original token from a given hidden state), showing that token identifiability rate decreases with depth, and that the cosine distance is more effective to
recover tokens than the $l_2$ distance. Moreover, experimental results indicate that the identity of tokens transforms as they pass through the various layers of the model; consequently, the fact that a hidden state is still traceable to the initial token does not necessarily imply that the embedding still represents the same information as the input. \citet{bert_linguistic} also analyze how information flows through BERT's \cite{BERT} layers, discovering good encoding of token positional information on BERT's lower layers, which transforms to a more hierarchically-oriented encoding on higher layers. These works are aligned with the findings of \citet{layer_importance}, which highlight the critical role played by the first layers in model performance. 

In this work we aim to shed light on the explainable role of the embedding layer with a \textbf{minimal perturbation ana\-lysis} across layers. This framework enables us to study the sensitivity of the embedding space $E$ to token perturbation. We present our findings as follows. Section \ref{sec:problem} introduces the optimization problem of minimal perturbation, followed by \autoref{sec:experiments} with our experiments. Within, section \ref{sec:frequency} we analyze which tokens, in terms of frequency, are responsible for minimal shifts on the embedding space. In particular, we determine whether there are discrepancies in terms of frequency when solving the optimization problem with various norms ($l_1,l_2,l_{\infty}$). Section \ref{sec:commonness} relates token commonness with their impact on $E$, concluding that rare words lead to larger shifts. Lastly, Section \ref{sec:layers} studies how these perturbations propagate through successive Transformer layers, and whether the original information of the input is preserved.

Our work is the first to combine token perturbations and shifts on $E$ to study the inner workings of the Transformer architecture. From this, our main insights are:

\begin{itemize}
    \item We investigate which tokens produce minimal shifts in the embedding space $E$ and the impact of the norm in selecting this minimal shift.
    \item We relate token commonness and their shift in $E$, concluding that rare words lead to larger perturbations.
    \item We analyze how information flows across layers from a perturbation and identification perspective. We assess the assumption that first layers can be used as proxies for explanations, with our experiments showing that the input evolves after the first layers. 
\end{itemize}

\section{Problem Definition} \label{sec:problem}
Understanding how information is represented and propagated through the internal layers of the Transformer remains a key challenge in model interpretability. Recent surveys \cite{survey_expl_transf,explainability_LLM,survey_LLM} emphasize that, although many interpretability approaches have been proposed, the embedding space has not been thoroughly examined.

In this section, we introduce the framework of minimal 
token perturbations. Formally, given an input sequence $x$, tokenized as
\[
\text{Tok}_1,\text{Tok}_2\ldots, \text{Tok}_{d_s}
\]
where $d_s$ denotes the sequence length, our goal is to analyze the effect of perturbations on $x$ and how they propagate to $E$ and the hidden states of the model. Specifically, for some index $i \in [d_s] = \{1, \ldots, d_s\}$, we replace $Tok_i$ with the most similar token in the token vocabulary $\mathcal{V}$ based on cosine similarity. Formally, the replacement token is selected as:
\begin{equation} \label{eq:tok_sim}
    \text{Tok}'_{i}=\argmax_{Tok \in \mathcal{V} \setminus\{Tok_i\}}\, \frac{E(\text{Tok}_i) \cdot E(\text{Tok})}{\|E(\text{Tok}_i)\| \|E(\text{Tok})\|}. 
\end{equation}
Given $\text{Tok}'_{i}$ then we construct $x'$, tokenized as  
\[
\text{Tok}_1,\text{Tok}_2\ldots, \text{Tok}_{i-1}, \text{Tok}'_{i}, \text{Tok}_{i+1}, \ldots, \text{Tok}_{d_s},
\]
which is $x$ with the replacement $\text{Tok}_i \mapsto \text{Tok}'_i$. Lastly, our goal is to compute $||E(x) - E(x')||$, namely the shift produced in the embedding space.

% \begin{equation} \label{eq:min_problem} 
%     \tilde{x} = \argmin_{\tilde{x} \in \mathcal{N}_x\setminus\{x\}} ||E(x) - E(\tilde{x})||
% \end{equation} 
% where $\mathcal{N}_x$ represents a suitable neighborhood of $x$.
% Our empirical analysis reveals that the minimum in equation \eqref{eq:min_problem} is typically achieved by altering a single token $Tok_i$
% Through a brute-force search over the vocabulary, we identify the token substitutions that result in the smallest shifts in the embedding space. In the following pages, when talking of solutions to eq. \eqref{eq:min_problem}, we improperly refer to the token to be altered to obtain $\tilde{x}$.

\section{Experiments and Discussion} \label{sec:experiments}
In the following experiments, we employ a BERT model \cite{BERT} fine-tuned on the IMDb Movie Reviews dataset \cite{imdb} from Hugging Face\footnote{
    https://huggingface.co/philipobiorah/bert-imdb-model
}. 
The test dataset consists of $25,000$ movie reviews, labeled as either positive $(1)$ or negative $(0)$. In particular, our analysis focuses on the encoder block of the architecture, which consists of $12$ attention layers mixed with residual connections and linear layers.

All experiments were run on Google Colab with the GPU T4 environment and can be executed in less than $10$min.

\subsection{Frequency Analysis of Minimal Perturbations} \label{sec:frequency}
\begin{figure}[!h]
    \centering
    \includegraphics[width=\columnwidth]{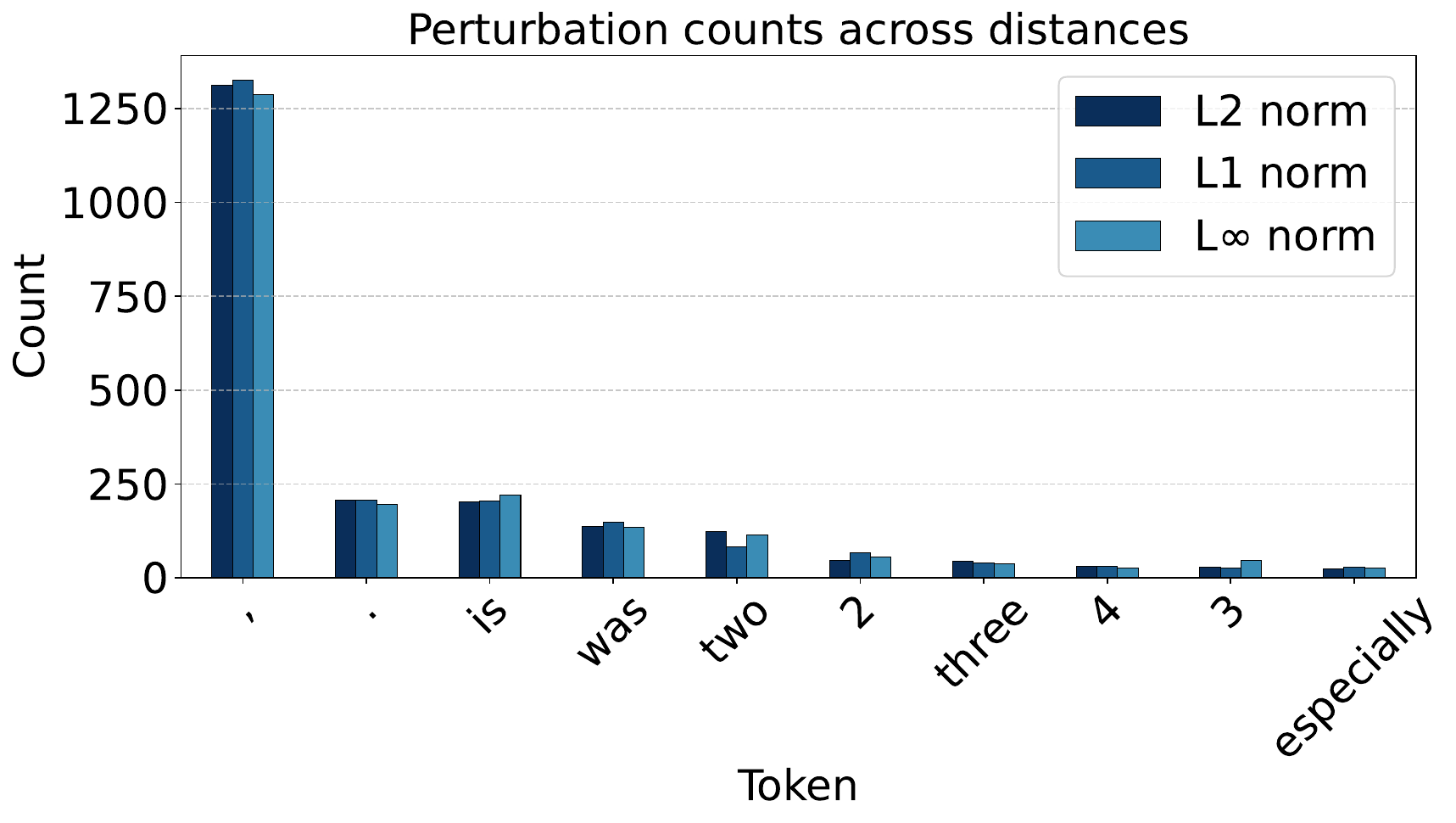}
     \caption{Tokens most frequently identified as minimally perturbing: we count how often each token appears among the top-5 least impactful substitutions (i.e., those inducing minimal shift in $E$) across $500$ sequences, separately for $l_1$, $l_2$, and $l_\infty$ norms.}
    \label{fig:counts}
\end{figure}

In this section, we investigate which parts of an input sequence are least sensitive to minimal perturbations. Given a sentence $x$, we identify the token that, when replaced with its nearest alternative (in cosine distance), results in the smallest embedding shift $||E(x) - E(x')||$. This analysis is repeated under different norms used to compute the shift, namely $l_1$, $l_2$, and $l_\infty$
For each of the $500$ sequences, we identify the \textbf{five tokens of the sequence} whose replacements lead to the smallest perturbation in terms of $||E(x)-E(x')||$.
\autoref{fig:counts} reports the distribution of replacements yielding minimal shifts for the $l_1,l_2, $, and $l_{\infty}$ norms. 

To assess the consistency of perturbation behavior across norms, we compute both the Pearson and Spearman correlations over substitution frequency profiles. The Pearson correlation $r$ shows very high agreement:
\[
r_{l_2,l_1}=0.99, \quad r_{l_2,l_{\infty}}=0.99, \quad r_{l_1,l_{\infty}}=0.99.
\]

This indicates that the norms largely agree on which token yields to minimal shifts. Similarly, the Spearman rank correlation, computed on the previous vectors but restricted to the top-20 entries (which account for ~$90\%$ of replacements) confirms high consistency:
\[
\rho_{\text{$l_2,l_1$}} = 0.96, \quad
\rho_{\text{$l_2,l_{\infty}$}} = 0.92, \quad
\rho_{\text{$l_1,l_{\infty}$}} = 0.95.
\]
It is worth noting that this high agreement is observed in the top-ranked region of the distribution. If the rank correlation were computed over the full list of substitutions, the value would likely be lower. This is primarily due to high variability in the tail, where low-frequency substitutions (often appearing only once among $500$ input sequences) have little statistical weight, yet contribute to rank noise. These rare replacements—though technically part of the solution to the perturbation minimization problem—are more sensitive to randomness and are less robust across distance metrics.

This empirical stability across norms aligns with the standard theoretical result that, in finite-dimensional vector spaces, all norms are topologically equivalent \cite{func_analysis}. As a result, optimization problems based on distance minimization—such as $||E(x)-E(x')||$—tend to exhibit similar behavior across norms, especially when the perturbations are constrained to lie within a compact and semantically meaningful region of the embedding space.

Finally, \autoref{fig:counts} suggests that the tokens producing the smallest changes in the embedding space are predominantly punctuation marks and numbers. First of all, these tokens are common (and hence, loaded with less semantic value, as discussed by \citet{word_length}). For instance, period ('.') and comma (',') have frequencies of $4.20\%$ and $3.54\%$, respectively. Moreover, considering that our BERT-based model is trained for sentiment analysis on movie reviews, this outcome is intuitive: replacing a comma, a period, or a number is unlikely to alter the semantic meaning in a way that significantly affects the model’s output.

\subsection{Perturbation Analysis based on Token Commonness} \label{sec:commonness}
We now investigate the relationship between word commonness and the shift in $E$ using the $l_2$ norm when altering such tokens. 
To quantify commonness, we first compute token frequencies from the corpus and apply a log transformation using:
\[
\text{freq}_{log}(w) := \log(1+\text{freq}(w))\,,
\]
where $\text{freq}(w)$ is the raw frequency count of token $w$. Next, we normalize these values using min-max normalization:
\[
\text{commonness}(w) = \frac{\text{freq}_{\log}(w) - \min_{w'} \text{freq}_{\log}(w')}{\max_{w'} \text{freq}_{\log}(w') - \min_{w'}\text{freq}_{\log}(w')}\,.
\]
We partition the resulting commonness scores into 10 equal-sized bins and select $50$ distinct sentences per bin (i.e., sentences containing tokens falling within that commonness interval).\footnote{Note that higher commonness bins contains fewer samples, because we associate one sentence to each token and these bins contain less than $50$ unique tokens.}

\begin{figure}[!h]
    \centering
    \includegraphics[width=\columnwidth]{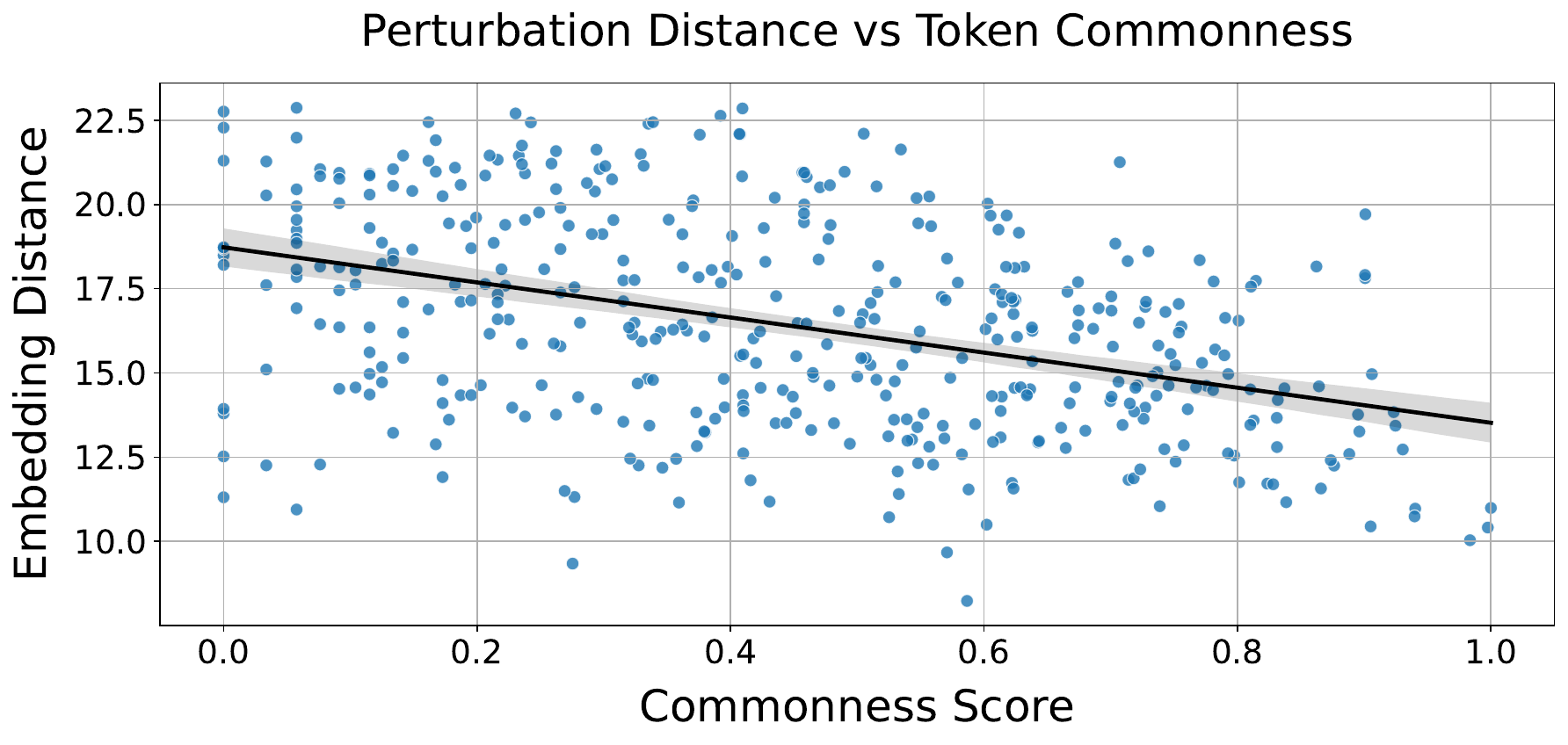}
    \caption{Analysis of embedding distances when perturbing tokens with specific commonness values. Linear regression with $95\%$ confidence interval, showing a significant downward trend.}
    \label{fig:rarity_vs_E}
\end{figure}

The plot in Figure \ref{fig:rarity_vs_E} relates commonness score against embedding distance, revealing a decreasing trend: rarer words induce a greater shift in the embedding space, compared to more common ones. The trend is confirmed by the regression line with a tight $95\%$ confidence interval.

A possible explanation for this behavior is the following. In our framework, each token in a sequence is replaced with its closest counterpart in the embedding space according to cosine similarity. Common tokens are more likely to have semantically similar neighbors, leading to replacements that preserve the overall meaning and structure of the sequence, and thus result in smaller shifts in the embedding space. Conversely, rare tokens tend to be underrepresented both in the vocabulary and in the learned embedding space. As a result, their replacements are less likely to retain the original semantics, producing a larger perturbation. Additionally, common tokens—such as punctuation or stopwords—carry limited semantic load, while rare tokens often play a more central role in sentence meaning. Perturbing them is therefore more likely to cause substantial changes in the representation. These conclusions complement the findings of \citet{word_length}, who showed that rare words tend to carry more information per token and are aligned with the observations of \citet{rare_word_problem}, which highlighted that rare words are often underrepresented (in the sense that they do not have their own embeddings) in the embeddings of BERT-like models.

\subsection{Propagation Across Layers} \label{sec:layers}
\begin{figure}[!h]
    \centering
    \includegraphics[width=\columnwidth]{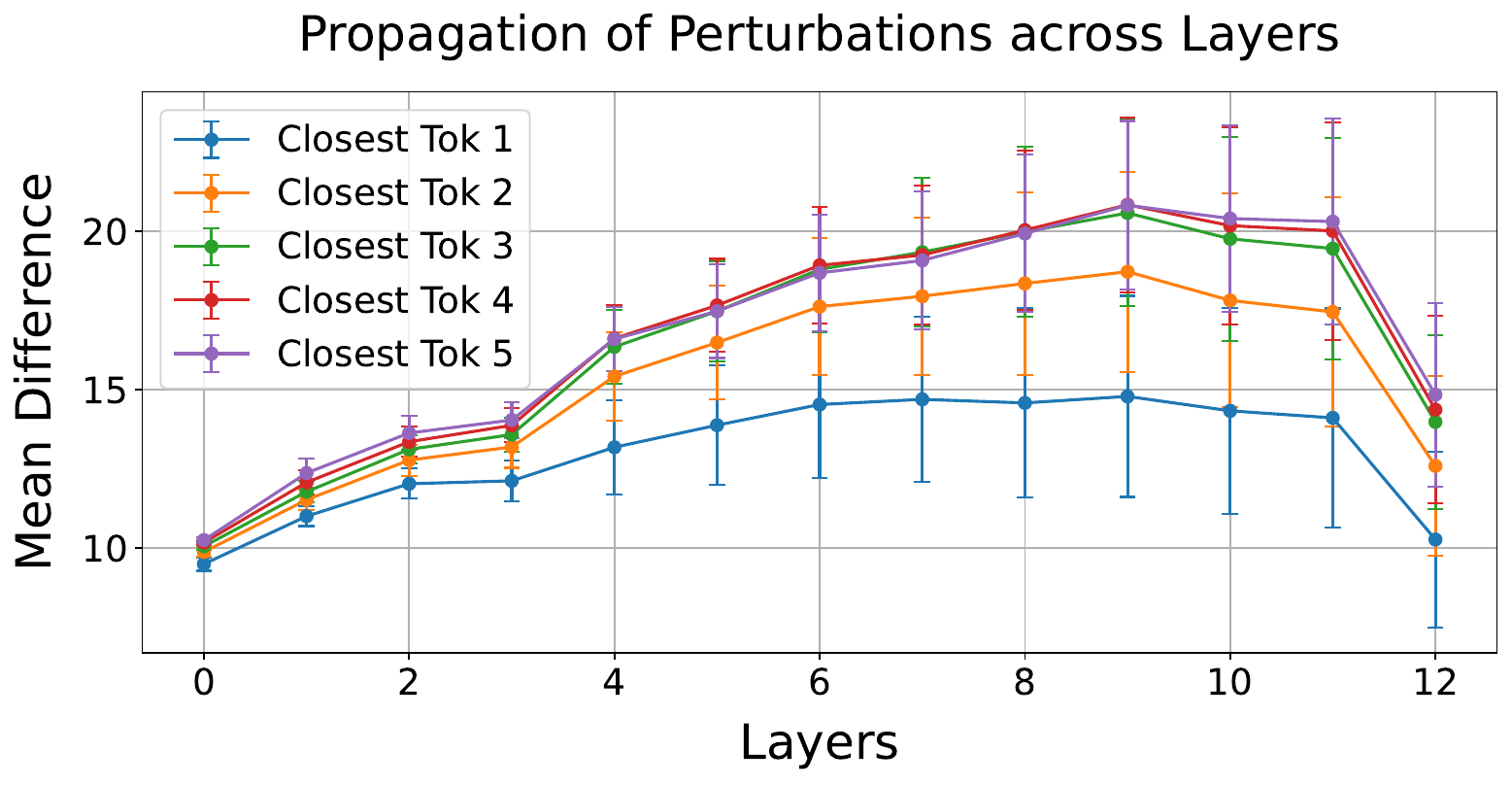}
    \caption{The propagation across layers of the top-5 least impactful substitutions for $500$ instances with $95\%$ confidence interval for the mean value. There is an overall increasing trend, except for the last layer.}
    \label{fig:layer_propagation}
\end{figure}

We recall that the BERT model under analysis consists of $12$ layers on top of the embedding layer; in this section, we aim to grasp how minimal perturbations propagate throughout the various layers of the encoder. In particular, we extract the top five tokens of the sentence whose replacement leads to the smallest perturbation for $500$ instances and compute the distance between the perturbed and original sentence across the hidden states using the $l_2$ norm. Interestingly, \autoref{fig:layer_propagation} shows that mean difference increases as we enter deeper layers, except for the last one. This inversion can be explained by the fine-tuning process: while earlier layers retain general-purpose representations, the final layer is typically adapted for the downstream classification task, behaving more like a task-specific classifier rather than part of the general encoding pipeline.
Moreover, in terms of trend, we observe that the first two tokens lead to smaller shifts in the hidden states, while the rest are almost indistinguishable. This reflects the fact that the first two tokens are indeed the most similar in terms of meaning and semantics w.r.t the token to be changed. Lastly, as we enter deeper layers, variance increases. This phenomenon supports the choice of using the first layers as proxies for explanations since they appear to be more robust to perturbations.

In light of the above, we question whether information is preserved across layers. In particular, \citet{tok_identifiability} discuss the identifiability of the original tokens across layers. They showed how the original tokens can be recovered employing MLPs and naive classifiers. Naive classifiers recover the original tokens across hidden states by retrieving the closest token in the embedding space in terms of cosine distance. They reason that as the information passes through more attention layers, the tokens become more intermixed, diminishing token identifiability. This stems from the fine-tuning, reflected as well on \autoref{fig:layer_propagation}. The last layers have a distinct behavior more specific to the task, employing a less identifiable and less interpretable encoding. That is, the embedding information becomes less relevant for the last hidden states.

These findings are aligned with the common assumption that first layers can be considered a proxy for explanation \cite{survey_expl_transf}. They also confirm the results of \citet{layer_importance}, who argue that the first layers of a network better preserve information and relevance.

In order to illustrate a fine-grained perspective of how each token embedding evolves across attention layers, we list the set of closest tokens in terms of cosine distance for each token position in the hidden state. Analogous to \eqref{eq:tok_sim}, we denote the vector corresponding to the token at position $j$, at hidden state $h_i$ as $h_{i,j}$. Then its closest token is:

\[
\argmax_{\text{Tok}\in\mathcal{V}} \frac{h_{i,j} \cdot E(\text{Tok})}{\|h_{i,j}\| \|E(\text{Tok})\|}. 
\]

For a specific instance from the test set:

\begin{quote}
    ``[CLS] could someone please explain to me the reason for making this movie? sad is about all i can say ; this movie took absolutely no direction and wound up with me shaking my head. what an awful waste of two hours. noth should be ashamed of himself for taking money for this piece of [SEP]"
\end{quote}

\noindent
The last hidden state closest-token association retrieves:

\begin{quote}
    ``planned ⟩ someone please [CLS] humor [CLS] [CLS] [CLS] is [SEP] scenery [SEP] [CLS] productive been chilean entire scenery stu done [SEP] scenery [SEP] scenery scenerycuit [CLS] scenery scenery scenery planned scenery courses scenery mouth sis scenery scenery scenery scenery scenery scenery scenery siskill [CLS] should be [SEP]ega planned is is money is planned scenery been planned"
\end{quote}

Now, if we focus on a single token and list its top-5 closest tokens across hidden states, we obtain how each $h_{i,j}$ evolves in relation to the embedding. For instance, in the case of ``ashamed" in the previous example:

\begin{quote}
 `ashamed embarrassed proud shame afraid', $E$ \\
 `ashamed embarrassed shame proud afraid', $h_1$ \\
 `ashamed embarrassed shame proud sorry',  $h_2$ \\
 `ashamed embarrassed shame sorry proud', $h_3$ \\
 `ashamed embarrassed shame sorry proud', $h_4$ \\
 `ashamed embarrassed shame sorry proud', $h_5$ \\
 `ashamed shame embarrassed sorry embarrassment', $h_6$ \\
 `ashamed embarrassed shame bad sorry', $h_7$ \\
 `ashamed shame embarrassed [SEP] bad', $h_8$ \\
 `ashamed [SEP] shame embarrassed bad', $h_9$ \\
 `[SEP] ashamed shame poor embarrassed', $h_{10}$ \\
 `[SEP] [CLS] later is and', $h_{11}$ \\
 '[SEP] 2005 courses 1985 productive' $h_{12}$ \\
\end{quote}

First layers contextualize each $h_{i,j}$, while deeper layers ($i\ge8$), become less identifiable and thus cannot be interpreted in terms of the embedding.  

\section{Limitations and Future Work}
While our framework provides valuable insights for BERT interpretability, several limitations should be acknowledged. Firstly, our definition of perturbation is limited to single-token substitutions based on cosine similarity. While this choice prevents from incurring into OoD behavior, more complex perturbations that might reveal additional properties of the model are left out. Secondly, our experiments focus exclusively on a BERT-based model trained on a sentiment analysis task. As a result, the generality of our findings to other architectures or tasks involving more structured outputs (e.g., question answering) remains to be studied.

Future work could address these limitations in several directions. Extending the perturbation framework to allow different kinds of perturbations (for instance, permutations or token shift) could provide a deeper understanding of the embedding space and its potential explanatory role. Moreover, exploring different tasks such as translation or summarization may enrich the knowledge and interpretability of these architectures. Indeed, we expect that tokens yielding to minimal perturbations are dependent on the specific task, as a result of the interaction between the type of model output and the semantic-syntactic value of the tokens.

\section{Conclusions}

In this work, we explored which tokens are responsible for minimal shifts in the embedding space of Transformers. Our analysis suggests that common tokens induce smaller embedding changes, highlighting the reduced semantic load carried by frequent tokens. Furthermore, we observe that perturbations have an increasing impact as information passes through deeper layers, thereby confirming the common assumption that early layers preserve the input more faithfully. These results align with prior findings on token rarity and identifiability, and support the use of early representations as interpretable proxies. In conclusion, our findings stress the important role played by the embedding space $E$ and the explanatory role of token perturbations.

\newpage
\section*{Acknowledgements}
The research leading to these results has received funding from the Horizon Europe Programme under the Horizon Europe Programme under the AI4DEBUNK Project (https://www.ai4debunk.eu), grant agreement num. 101135757. Additionally, this work has been partially supported by  PID2019-107255GB-C21  funded by MCIN/AEI/10.13039/501100011033 (CAPSUL-IA) and JDC2022-050313-I  funded by MCIN/AEI/10.13039/501100011033al by European Union NextGenerationEU/PRTR and the “Generación D” initiative, Red.es, Ministerio para la Transformación Digital y de la Función Pública, for talent attraction (C005/24-ED CV1), funded by the European Union NextGenerationEU funds, through PRTR.

\bibliographystyle{named}
\bibliography{ijcai25}
\nocite{*}

\end{document}